\documentclass[runningheads]{llncs}
\usepackage{amsmath}
\usepackage{amssymb}
\usepackage{color} 
\usepackage{subfig}
\usepackage{xstring}
\usepackage{url}

\usepackage{pdflscape}
\usepackage{afterpage}
\usepackage{capt-of}%
\usepackage{nicematrix}
\usepackage{multirow}

\usepackage{catchfilebetweentags}
\usepackage{booktabs}
\usepackage{todonotes}
\usepackage{wrapfig}

\newcommand{\alt}[3]{%
    \IfEqCase{#1}{%
        {1}{#2}%
        {2}{#3}%
    }%
}%

\definecolor{cen}{rgb}{0.12156862745098039, 0.4666666666666667, 0.7058823529411765}
\definecolor{dec}{rgb}{1.0, 0.4980392156862745, 0.054901960784313725}
\definecolor{cc}{rgb}{0.17254901960784313, 0.6274509803921569, 0.17254901960784313}
\definecolor{cd}{rgb}{0.8392156862745098, 0.15294117647058825, 0.1568627450980392}
\definecolor{dd}{rgb}{0.5490196078431373, 0.33725490196078434, 0.29411764705882354}
\definecolor{cenTL}{rgb}{0.6823529411764706, 0.7803921568627451, 0.9098039215686274}
\definecolor{decTL}{rgb}{1.0, 0.7333333333333333, 0.47058823529411764}

\usepackage{graphicx}
\begin{document}
\title{Hierarchical Decentralized Deep Reinforcement Learning Architecture for a Simulated Four-Legged Agent}
\titlerunning{Hierarchical Decentralized Deep Reinforcement Learning Architecture}
%
\author{Wadhah Zai El Amri\inst{1}\orcidID{0000-0002-0238-4437}\and
Luca Hermes\inst{1}\orcidID{0000-0002-7568-7981} \and
Malte Schilling\inst{1}\orcidID{0000-0002-0849-483X}}
\authorrunning{W. Zai El Amri et al.}
%
\institute{Machine Learning Group, Bielefeld University
\email{\{wzaielamri,lhermes,mschilli\}@techfak.uni-bielefeld.de}}
\maketitle              
\begin{abstract}
Legged locomotion is widespread in nature and has inspired the design of current robots. 
The controller of these legged robots is often realized as one centralized instance. However, in nature, control of movement happens in a hierarchical and decentralized fashion.
Introducing these biological design principles into robotic control systems has motivated this work. We  tackle the question whether decentralized and hierarchical control is beneficial for legged robots and present a novel decentral, hierarchical architecture to control a simulated legged agent.
Three different tasks varying in complexity are designed to benchmark five architectures (centralized, decentralized, hierarchical and two different combinations of hierarchical decentralized architectures). The results demonstrate that decentralizing the different levels of the hierarchical architectures facilitates learning of the agent, ensures more energy efficient movements as well as robustness towards new unseen environments. Furthermore, this comparison sheds light on the importance of modularity in hierarchical architectures to solve complex goal-directed tasks. We provide an open-source code implementation of our architecture (\url{https://github.com/wzaielamri/hddrl}).
\end{abstract}
\section{Introduction}
Legged locomotion has been widely used in current mobile robots~\cite{hutter2016,schneider2014} as it provides a high degree of mobility and is adequate for various types of terrains. This makes legged robots useful for different application scenarios, from assisting humans in their daily lives to rescue missions in dangerous situations. Researchers have developed several ideas and algorithms to design and program legged robots. 
One important inspiration for the field comes from biology as walking animals show high stability, agility, and adaptability. 
Insects provide one example: They can climb and move in nature while relying on comparably simple control structures. At the same time, they only have limited knowledge of their surroundings. 
Still, adaptive behavior emerges and enables these animals to move successfully and efficiently. Such capabilities are desirable for today's robots. 
A drawback is that the transfer of biological principles usually requires detailed knowledge and explicit setup or programming of control systems. This is difficult to scale towards real world settings covering a wide range of scenarios.

Another approach to robotic control is based on machine learning. 
While common reinforcement learning approaches have been successfully applied in robot locomotion, these systems often lack generalizability towards multiple tasks~\cite{frans2017}. In particular, it has shown to be non-trivial to design reward functions for each task. Ideally, it would be sufficient to simply reward completion of the full task, e.g. reaching a goal.
But it is challenging for an algorithm to learn goal-directed behavior from such sparse feedback. Learning from sparse rewards requires temporal abstraction that contributes to solve multiple tasks and adapt to different environments~\cite{kulkarni2016}. 
Animals are successful at such tasks as they rely on a hierarchical organization into modules in their control systems~\cite{merel2019}. By splitting the locomotion system into different hierarchical levels (vertical levels), it becomes possible to coordinate, acquire and use knowledge gained in previous tasks. Reinforcement learning algorithms have adopted this notion of hierarchical organization~\cite{merel2019} in the form of hierarchical networks and applying transfer learning where primitive behaviors are stored and can be reused in new tasks.

Recently, the idea of multiple decentralized modules (horizontal modularization) running concurrently to ensure high-speed control and fast learning has been introduced to Reinforcement Learning~\cite{schilling2021}. Such a solution can be found in nature, e.g., in insect locomotion it is assumed that each leg is controlled by a single module that adapts to local disturbances based only on local sensory feedback~\cite{schilling2013}. 
Both these types of modularization---into a hierarchical and decentralized organization---have been used separately in state-of-the-art legged mobile robots. In this article, we combine these concepts in a hierarchical decentralized architecture and consider how this helps learning behavior as well as how this positively influences behavior, in particular generalization towards novel tasks and more efficient behavior. The paper is structured as follows: General concepts and ideas are discussed in related work to address the aforementioned challenges. Sec.~3 introduces the methodological foundations of this work and addresses the experimental setup designed to deal with the research questions. Sec.~4 shows the main results of the conducted experiments and the various investigated details, such as the emerged new learning paradigms out of this novel concept. Sec.~5 discusses further aspects and concludes this paper. 

\section{Related Work}
%
%
As we aim to take inspiration from animal control, we focus on two key aspects: decentralization and hierarchical organization. Merel et al.~\cite{merel2019} provided an analysis of the mammalian hierarchical architecture and gave several design ideas for hierarchical motor control. Quite a number of scientists have already taken advantage of such an hierarchical architecture for legged locomotion. For example, Heess et al.~\cite{heess2016} introduced a novel hierarchical deep reinforcement learning (DRL) architecture. The controller consists of two policies, a high-level and a low-level policy. The low-level controller (LLC) is updated at a high execution frequency while the high-level controller (HLC) operates on a low frequency. 
The HLC has access to the entire observation space including task-related information such as the goal position. However, the LLC has access only to local inputs corresponding to the agent's state and latent information originating from the HLC. The learned low-level motor behaviors can be transferred to various other, more complex tasks---such as navigation---while getting a sparse reward \cite{schilling_approach_2018}.
Other approaches, e.g.~\cite{li2020}, use a hybrid-learning model. In their approach \cite{li2020}, the HLC is model-based, and it selects the appropriate policy to use. Whereas the LLC is trained with an off-policy learning-based method. 
In \cite{azayev2020} a simulated robot learned to walk through environments that differed in terrain or had obstacles. LLCs were trained separately for a specific environment, and a HLC was trained afterwards to select the suitable LLC for the current environment.


%
%
Decentralization has been identified as another key principle in biological motor control~\cite{schilling2013}. It has already been applied in DRL in \cite{schilling2021,schilling2020decentralized}. 
They proposed a way to take advantage of a decentralized architecture in learning of legged locomotion. DRL was used to train four independent policies for a four-legged agent, one for each leg. 
They further 
regulated the information flow for the individual (leg) policies, 
e.g., instead of directing all available input information to the controller, the information was split, and each of the four agents received only local information relevant to them. 
The results of the paper clearly demonstrated that the decentralized architecture provides an enhancement of the speed of learning by reaching high performance values compared to centralized architectures. In addition, the learning process is more robust to new unpredictable environment sets, such as changes in the flatness of the terrain. A different approach to decentralization was proposed by Huang et al.\cite{huang2020}. They trained shared modular policies, each responsible for an actuator of the agent. 
With the help of the messaging between the reusable policies, a communication between the different modules emerges, and this communication ensures a stable movement, which is robust to various changes in the physiology of the agent. 
But this approach took a long period of time and requires high computational resources to obtain a trained shared policy.
Both approaches used a simple 1-D forward movement as a task 
where the agent received a reward for the speed of movement in the x-direction with other penalties such as contact or control punishments. However, it is important to note that the decentralized architectures in these works are not tested on more complex tasks. 
%
%
In this work, we introduce novel architectures that attempt to combine both approaches and use DRL to train the different policies when applied to more complex tasks.

\section{Methods}
\label{sec:methods}
The aim of this study is to compare how decentralization and hierarchical organization of control structures influence adaptive behavior and learning of behavior. In particular, different architectures for walking of a four-legged simulated agent are evaluated with respect to how fast stable behavior emerges, how robust and efficient learned behavior is, and, finally, how well an architecture supports transfer learning, i.e. being applied in a different setting. We experiment using central and decentral walking policies in hierarchical and non-hierarchical setups. We start by introducing these principles and then describe the specific architectures.

Current reinforcement learning (RL) approaches are usually based on central policies $\pi_{central}$ as the most straight forward approach. Such a central policy (equation \ref{equ:policy}) consists of a single module that receives the full observation vector $\mathbf{s}_t$ at time step $t$ and outputs control signals $\mathbf{a}_t$ for each degree of freedom (DOF). In contrast, a decentral control system consists of separate sub-controllers $\pi^{l}_{\text{decentral}}$. Each sub-controller only outputs the control signal for the DOFs for its assigned leg $l$ and usually receives only local observations $\mathbf{s}^l_t$, for example, only from the respective leg considering control of a walking agent. As mentioned, decentralization can be found throughout biological control as, for example, it is required to compensate for sensory delays. In RL, usually decentralization is abstracted away and not considered because it is only assumed as an implementation detail. In contrast, our study aims to analyze the contribution of decentralization as we assume that decentralization itself has positive effects.
In the case of the four-legged agent used here (see sect. \ref{subsect:setup}), for a decentralized approach the four legs require four sub-controllers that are trained concurrently. Note, that it is possible to choose other patterns of decentralization. Instead of per-leg controllers, per-joint controllers would be an option, too, which is not explored in this work.

\begin{equation}
\label{equ:policy}
\pi_{\text{central}}(\mathbf{a}_t | \mathbf{s}_t), \quad\quad \pi^{l}_{\text{decentral}}(\mathbf{a}^l_t|\mathbf{s}^l_t)
\end{equation}

Decentralization introduces a form of modularization into multiple modules on the same level of abstraction that act concurrently (one controller for each leg, but these are equal in structure). A different form of modularization is introduced by an hierarchical structure. In a hierarchical structure higher levels usually drive lower level modules. This has the advantage that lower level modules can be learned as reusable building blocks that can be applied by higher levels in multiple tasks. Such hierarchical approaches facilitate transfer learning. 
A non-hierarchical setup---as the standard RL approach---consists of a single level of control, as observations are mapped onto actions by a policy neural network. In a hierarchical setup the controller is distributed onto two, or more, levels of a hierarchy. In our case, we use two levels of hierarchy and define our principles of hierarchy following Heess et al. \cite{heess2016}: Low-level controller (LLC) output actions at every time step $t$ and high-level controller (HLC) output a latent vector $\mathbf{m}_\tau$ every $\tau$ time steps (equation \ref{equ:hierarch}). This introduces a form of temporal abstraction. The LLC only gets local features $\mathbf{s}_t$---like the joint angles---as an input, and is in addition conditioned on the latent vector $\mathbf{m}_\tau$ from the HLC. This allows the HLC to modulate the LLC. The HLC only gets global features $\mathbf{\bar{s}}_\tau$, such as the angle between robot orientation and target position.

\begin{equation}
\label{equ:hierarch}
    \mathbf{m}_\tau = \pi_{HLC}(\mathbf{\bar{s}}_\tau), \quad \mathbf{a}_t = \pi_{LLC}(\mathbf{m}_\tau \parallel \mathbf{s}_{t})
\end{equation}



\begin{figure}[!tb]
    \centering
    \includegraphics[scale=0.2]{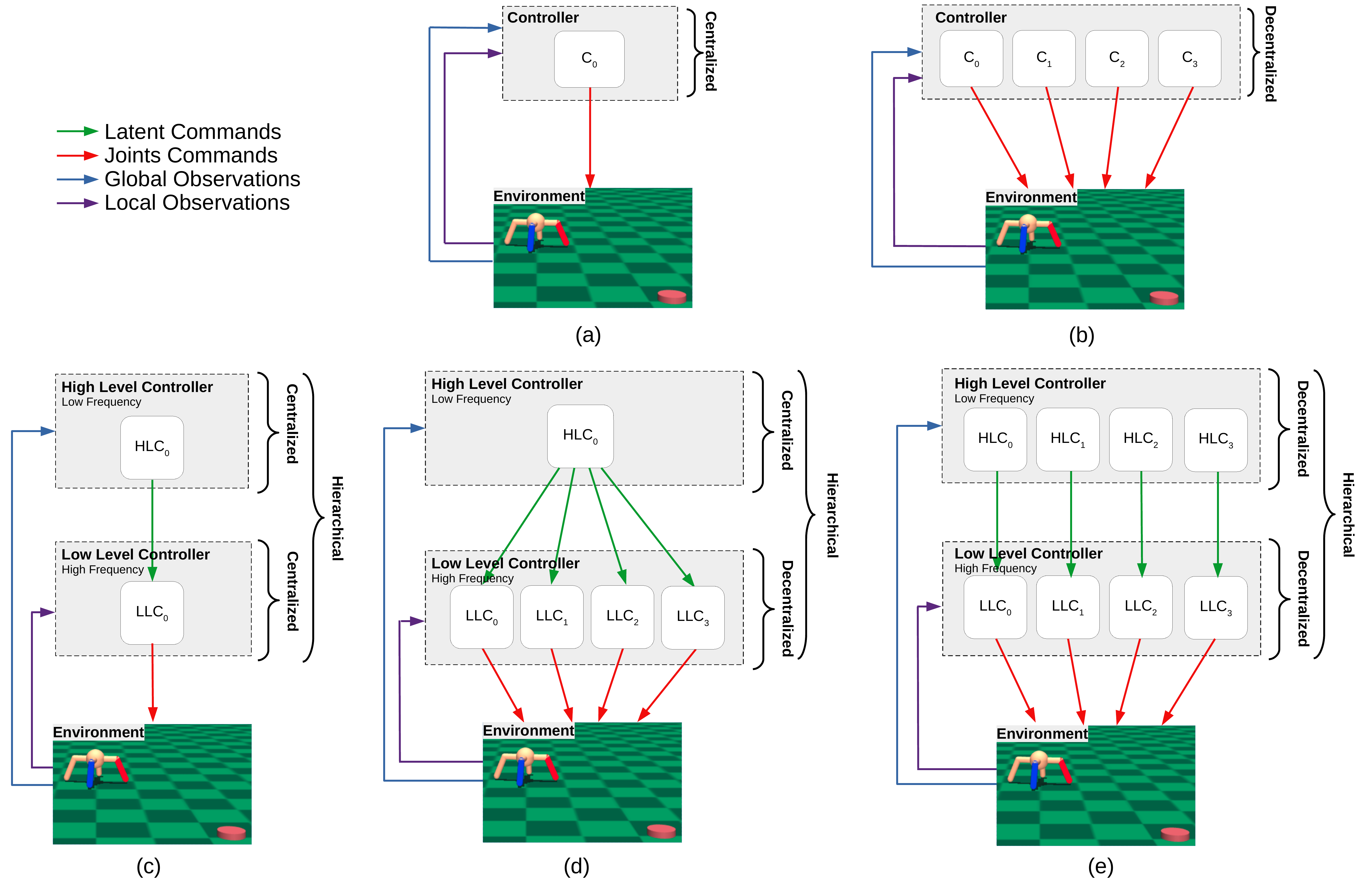}%
   
    \caption{The different architectures. (a): Centralized architecture with one controller ($C_0$, termed Cen). (b): Decentralized architecture with four different controllers ($C_0$, $C_1$, $C_2$, $C_3$, termed Dec). (c): Hierarchical architecture with 1 HLC and 1 LLC (each level consists of a single, centralized controller and there is no decentralization on each level of abstraction; therefore, this controller is abbreviated as C\_C). (d): Hierarchical decentralized architecture with 1 HLC and 4 LLCs (abbr. as C\_D as the higher level consists of a single, central module and the lower level is decentralized). (e): Hierarchical decentralized architecture with 4 HLCs and 4 LLCs (abbr. as D\_D).}%
    \label{fig:architectures}%
\end{figure}

We experiment with five different architectures: On the one hand, employing known architectures as centralized, decentralized, and an hierarchical approach. On the other hand, as the main novel contribution of this article, we test two combinations of hierarchical and decentralized architectures (Fig. \ref{fig:architectures} visualizes all five explored architectures). 
The central architecture (shown in Fig. \ref{fig:architectures} (a)) serves as the central baseline. The decentralized architecture (shown in Fig. \ref{fig:architectures} (b)) serves as the decentral baseline, replicating the architecture in \cite{schilling2021}. 
To investigate if decentralization benefits hierarchical controllers, we experiment with combinations of central and decentral policies on two different levels of hierarchy, as shown in Fig. \ref{fig:architectures} (c)-(e). 



\subsection{Experimental Setup}
\label{subsect:setup}
The experiments focus on learning locomotion tasks for a four-legged simulated agents. We used the MuJoCo physics simulator \cite{todorov2012} (v. 1.50.1.68) to run the experiments and adopt the modified Ant-v3 model from \cite{schilling2021}. This version of Ant was made heavier by a factor of ten to replicate the weight of a real robot. We used the RLlib framework (v. 1.0.1) \cite{liang2017} and applied PPO \cite{schulman2017} in an A2C setup \cite{konda1999} to optimize our policies. Actor and critic shared a similar architecture. Both were realized as \textit{tanh}-activated two hidden-layer neural networks with 64 neurons in each layer. Note that the last layer of the critic network consists of a single neuron---which estimates a value for a given state---and the last layer of the actor network consists of twice as many neurons as there are actuated DOFs to parameterize a multivariate Gaussian. As our third experiment involved more complex navigation, we added an LSTM cell with 64 
units prior to the output layer which should allow for memorizing the path through a maze.

\subsection{Tasks}
The different policies are evaluated on three tasks from the literature that also involve high-level decision-making in contrast to simply walking in a straight line. The first two tasks are taken from \cite{heess2016} and the last one from \cite{duan2016}. The observation-space changes slightly between the tasks to include target-related information. Table \ref{tab:task_observations} shows the observations for all tasks. 
As we are aiming for transfer towards more complex tasks, we follow a sequential approach of learning for the hierarchical architectures as often employed in transfer learning. First, the models are trained on the first task. Secondly, in the case of the hierarchical models, the parameters of the LLCs of the trained models are frozen. This means, the LLC is kept fixed and the HLC is trained from scratch for task two and three, respectively, for the hierarchical architectures. This ensures that the LLC learns a universal policy for general leg control. In the non-hierarchical approaches (centralized architecture and decentralized architecture, Fig. \ref{fig:architectures} (a) and (b)), as there is no distinction between high and low-level, these models are trained directly on tasks two and three and there is no freezing of parameters.

\begin{table}
\caption[Observation spaces of the three tasks: divided into global and local information.]{Observation spaces of the three tasks divided into global and local information; $d$ denotes the dimensionality of the corresponding vector.}
\label{tab:task_observations}  
\resizebox{\textwidth}{!}{
\begin{tabular}{p{1.7cm}p{6.3cm}cccc}
\hline\noalign{}
Group & Observation & d & First Task & Second Task & Third Task \\
\noalign{}\hline
\hline
Global: & Angle between target and robot’s ”north pole” & 1 & x & x & x \\
\cline{2-6}
& Robot’s position: x, y, z & 3 & & x & x \\
\cline{2-6}
& Torso’s orientation: quaternion & 4 & & x & x \\
\cline{2-6}
& Robot’s velocity: x, y, z & 3 & & x & x \\ 
\cline{2-6}
& Robot’s 
angular velocity: $\omega_x, \omega_y, \omega_z$ & 3 & & x & x \\
\hline
\hline
Local: & Joints angles & 8 & x & x & x \\
\cline{2-6}
& Joints angular velocities & 8 & x & x & x \\
\cline{2-6}
& Passive forces exerted on each joint & 8 & x & x & x \\
\cline{2-6}
& Last executed joint actions & 8 & x & x & x \\ \hline
Total dims: & & 46 & 33 & 46 & 46 \\
\noalign{}\hline
\end{tabular}
}
\end{table}

\noindent \textbf{Task 1:}~In the first task, the agent should reach a specific target location $\mathbf{g}$. The initial location of the target and the agent are randomly chosen, such that the distance between robot and target ranges from 0.5m to 5m. This task is considered solved if the agent reaches the target within $t_\text{max} = 300$ simulation steps. The target is reached if the robot is within in a 0.25 m radius around the target location. For the two non-hierarchical models as well as the LLC of the hierarchical models, the following equation is used to calculate the continuously provided reward at timestep $t$
\begin{align*}
r_1(t) &= \frac{v_g(t)}{N} - 0.05 \times \| \mathbf{a}(t) \|^2 + r_T(t)\\
r_T(t) &= \begin{cases}
    0.2 \times (t_{\text{max}} - t),  & \text{if target reached}\\
    0,                          & \text{otherwise}
\end{cases},
\end{align*}
where $v_g$ denotes the velocity towards the target, $N$ denotes the number of sub-controllers, i.e. $N = 1$ for the central policy and $N = 4$ for the decentral case, $\mathbf{a}$ denotes the action vector and $\mathbf{p}$ denotes the robot position. The HLCs in hierarchical policies receive the following reward:
\begin{align*}
\bar{r}_1(t) = \cos \Theta(t) \times \| \mathbf{p}(t) \|^2 + 0.1 \cdot r_T(t)
\end{align*}
where $\Theta$ denotes the angle between the orientation of the robot and the direction from robot to target.
\\

\noindent \textbf{Task 2:}~~The second task is again to reach a set goal location, but this time only a sparse reward is given. The policy only collects a reward at the end of an episode on reaching the goal and not after every simulation set. 
We apply the following equation to calculate the reward at timestep $t$
\begin{align*}
\bar{r}_2(t) = \begin{cases}
    \frac{1}{N} - \frac{t}{N \cdot t_\text{max}},  & \text{if target reached}\\
    0,                          & \text{otherwise}
\end{cases}.
\end{align*}
\\

\begin{figure}
    \centering
    \includegraphics[width=.28\linewidth]{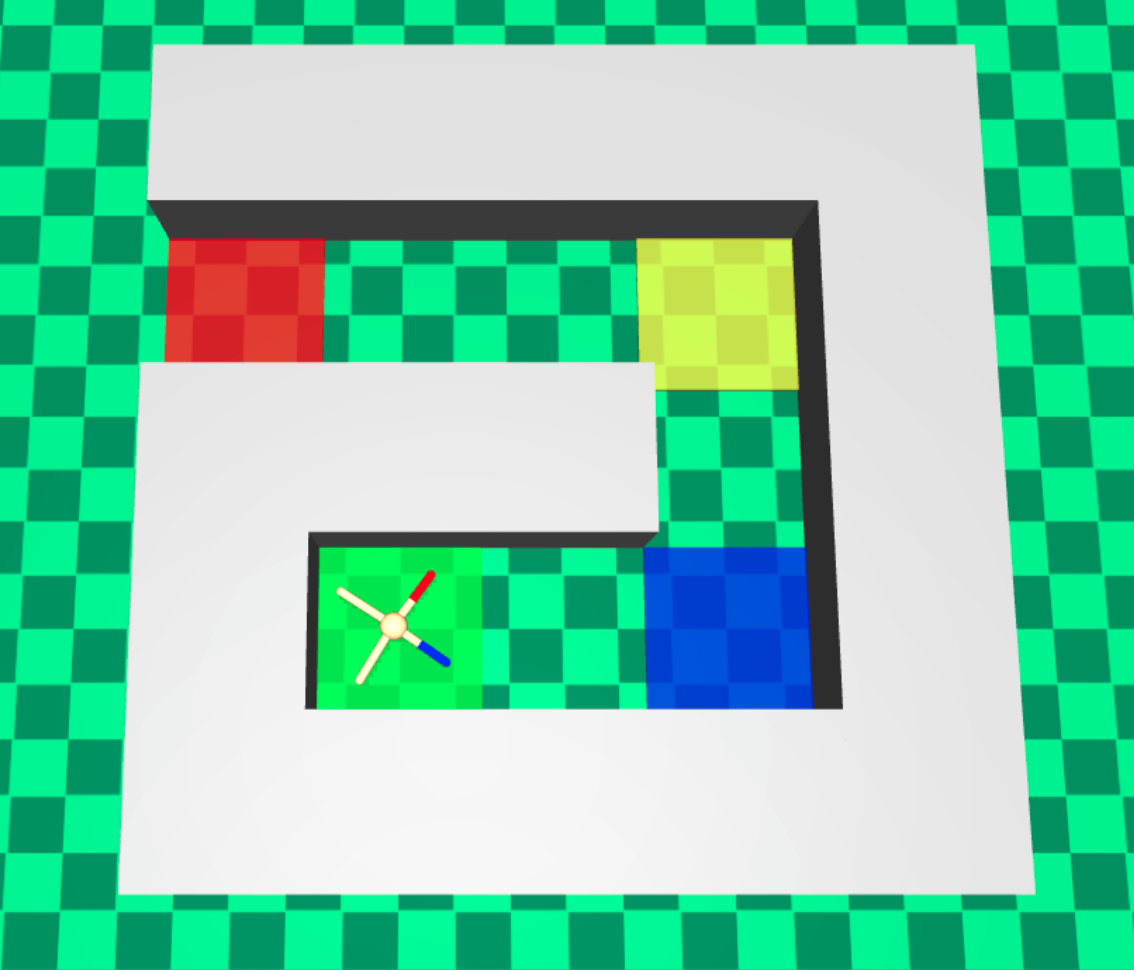}
    \caption[Maze environment of the third task.]{Maze environment of task three. The green square is the starting position, the blue, yellow, and red square represent the subgoals.}
    \label{fig:task3_new}     
\end{figure}

\noindent \textbf{Task 3:}~In the third task, the robot is put in a $15 \times 15\,m$ maze, as shown in Fig. \ref{fig:task3_new}. The robot always spawns in the green area and has to reach the red area within 1600 simulator steps. The initial orientation of the robot is randomized. This task is formulated in a sparse reward setting, but there are a couple of subgoals provided that should help guide the agent towards the goal. The agent gets as a reward 
\begin{align*}
\bar{r}_3(t) = \begin{cases}
    \frac{r_\text{subgoal}}{N},  & \text{upon first visit at subgoal}\\
    0,                           & \text{otherwise}
\end{cases},
\end{align*}
where $r_\text{subgoal}$ is 1, 2, or 3 for the blue, yellow, and red subgoal, respectively.
\\

\section{Results}
Three different tasks with increasing complexity have been studied. In each task, the five architectures introduced in Sec. \ref{sec:methods} were evaluated by running ten trials. During each trial, training ran for 40M simulation time steps. A single episode was terminated as soon as the goal condition was reached. 
We report the return accumulated over running an episode, the ratio of successful episodes and also power consumption of the agent as quantitative metrics. Throughout the result section, results will use abbreviations to represent task and architectures (first, providing the task number, followed by an underscore and the type of architecture; flat architectures (centralized, decentralized) are abbreviated as Cen and Dec, respectively, while for the hierarchical architectures it will be provided individually for the higher and lower level, if the policies are decentralized (D) or central (C) on that level, see Fig. \ref{fig:architectures} for all architectures; furthermore, we will use color coding throughout the results to distinguish architectures). 
The power consumption $P$ is calculated over time as the product of the torque acting on each joint along with its actual velocity, i.e. $P(t) = \sum_{\forall i} a_i \omega_i$, where $a_i$ is the action and $\omega_i$ the angular velocity of DOF $i$.
We report the return-specific results for hierarchical and non-hierarchical policies in separate plots as these differ in the reward functions used. Furthermore, we provide in addition the ratio of successful episodes as a fair measure for comparison.

In addition to providing training statistics, all agents were afterwards tested for 100 episodes. This test was repeated with three different randomly generated terrains. The terrains were generated using the algorithm for terrain generation from \cite{tassa2018}. It generates random height fields consisting of various superimposed sinusoidal shapes. The $smoothness$ parameter determines how uneven the terrain should be. This parameter ranges from 1.0 (flat terrain) to 0.0 (very uneven terrain). Three parameter values were tested in our experiments: flat terrain ($smoothness=1.0$), slightly uneven terrain ($smoothness=0.8$), and bumpy terrain ($smoothness=0.6$).

\begin{figure}[!tb]
    \centering
    \subfloat[\centering ]{{\includegraphics[scale=0.25]{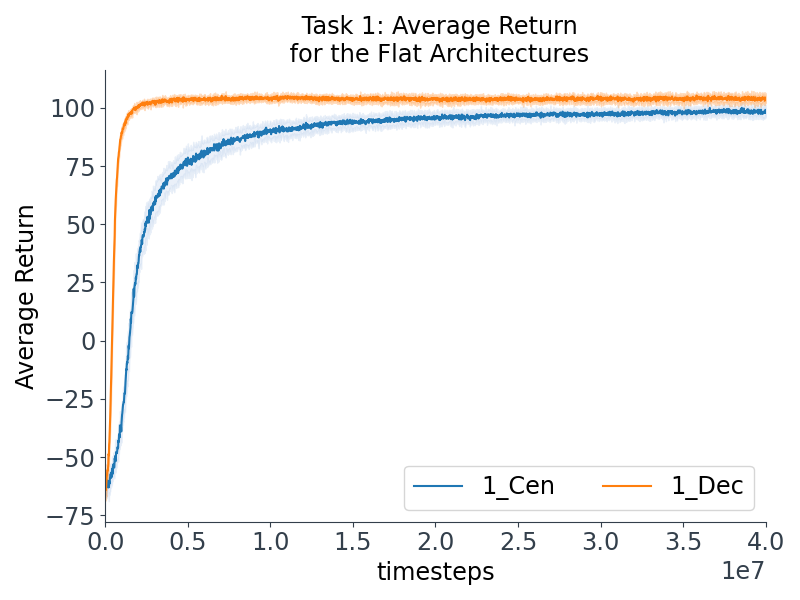} }}%
    \qquad
    \subfloat[\centering ]{{\includegraphics[scale=0.25]{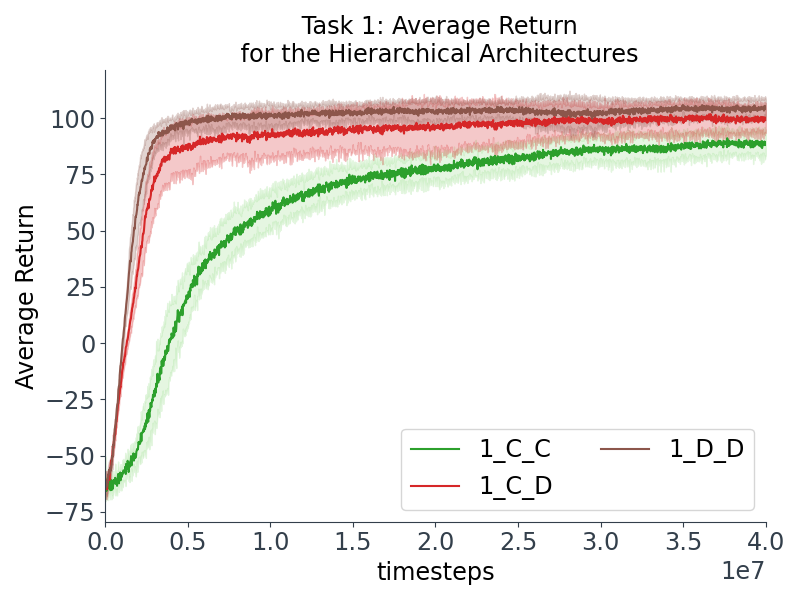} }}%
    \caption[Task 1: Average return of ten trials for the different architectures.]{Average return of ten trials for the different architectures during learning. (a) The average return of all ten trials for the flat architectures. The shaded area represents the limits of the standard deviation and the plot represents the mean value. (b) The average return of all ten trials for the hierarchical architectures. The shaded area represents the limits of the standard deviation and the plot represents the mean value.}%
    \label{fig:task1_averageReward}%
\end{figure}

\subsection{First Task: Navigate to a Specific Goal}
In the first task, the agent's goal was to reach a random target location and he was provided all the time with a continuous reward guiding him to the target location. Fig.~\ref{fig:task1_averageReward} (a), (b) show the development during training as the mean over all trials for the flat architectures and the mean return for the hierarchical architectures. The shaded areas show the standard deviation. 
It can be observed that the decentral architecture improved faster than the central architecture. This replicates the results of Schilling et al. \cite{schilling2021}, even for the more complicated task of walking towards a random target position. 
Fig. \ref{fig:task1_averageReward} (b) shows that the fully centralized architecture (\textcolor{cc}{1\_C\_C}) trained the slowest, whereas the fully decentralized architecture (\textcolor{dd}{1\_D\_D}) was learning the fastest. This indicates that the result of \cite{schilling2021} also appear to apply for hierarchical cases.

\begin{figure}[!b]
    \centering
    \includegraphics[scale=0.28]{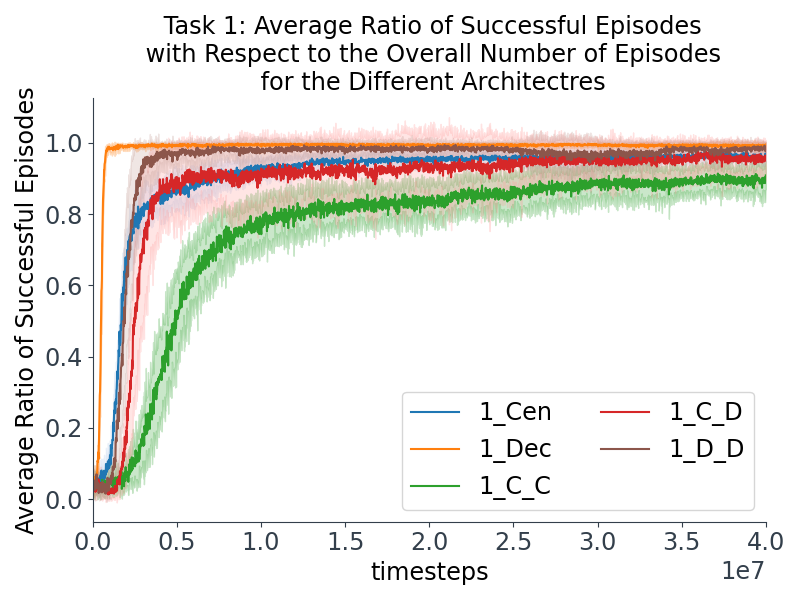}
    \caption[Task 1: Visualization of the ratio of successful episodes with respect to the overall number of episodes.]{Visualization of the ratio of successful episodes with respect to the overall number of episodes.}
\label{fig:task1_successEpisode}     
\end{figure}

Fig.~\ref{fig:task1_successEpisode} shows that agents with a flat decentralized architecture (\textcolor{dec}{1\_Dec}) learned the task the fastest and had a high success rate. The hierarchical architecture with decentralized HLCs and LLCs (\textcolor{dd}{1\_D\_D}) learned slightly slower. A possible explanation may be the hypothesis of Schilling et al. \cite{schilling2021} that a larger observation space requires longer training time. In this particular case, the observations space for the hierarchical policies was larger due to the introduction of the latent modulation vector from the HLC. The fully central hierarchical policy learned the task the slowest and only finished roughly 90\% of episodes successfully.

\begin{table}[!b]
\caption[Task 1: Results of the trained controllers in the deployment phase.]{Task 1: Results of the trained controllers after training. Given are mean values and standard deviation (in parenthesis) over 100 episodes for each of the ten trained seeds. Best values are highlighted (as the reward functions differ for the flat and hierarchical architectures, there is one highlight for each of these types (flat and hierarchical architectures). Therefore, these return values can not be compared directly, 
but we report this value for completeness. $P_{suc.}$ denotes the power consumption in successfully finished episodes.}
\label{tab:task1_deployment}
\resizebox{\linewidth}{!}{%
\begin{tabular}{@{\extracolsep{2mm}}l l lll l lll l lll}
\hline\noalign{}
Architecture & & \multicolumn{3}{c}{Flat terrain (1.0)} & & \multicolumn{3}{c}{Uneven terrain (0.8)} & & \multicolumn{3}{c}{Bumpy terrain (0.6)} \\
\cline{3-5}\cline{7-9}\cline{11-13}\noalign{}
(HLC + LLC)  & & Return\textsuperscript{*} & Ratio & $P_{suc.}$ & & Return\textsuperscript{*} & Ratio & $P_{suc.}$ & & Return\textsuperscript{*} & Ratio & $P_{suc.}$  \\
\noalign{}\toprule
\textcolor{cen}{Centralized}      & & 99.08 & 0.971 & 879.47 & & 97.16 & 0.965 & 867.22 & & 94.55 & 0.932 & 938.32\\
 & & ($$22.48) & & ($$364.65) && ($$23.57) & & ($$356.51) & & ($$25.36) & & ($$407.12)  \\

\hline
\textcolor{dec}{Decentralized}    & & 104.47 & \textbf{0.993} & 578.29 & & 103.12 & 0.988 & 585.96 & & 99.41 & \textbf{0.963} & 625.93\\
     & & ($$19.45) & & ($$250.50) && ($$19.94) & & ($$250.43) & & ($$21.86) & & ($$275.78) \\
\toprule
\textcolor{cc}{Cent. + Cent.}     & & 90.01 & 0.895 & 1240.55 & & 89.15 & 0.911 & 1276.92 & & 83.56 & 0.836 & 1355.97\\
      & &  ($$31.08) & & ($$577.86) & & ($$30.79) & & ($$623.84) & & ($$33.91) & & ($$668.56)  \\
\hline
\textcolor{cd}{Cent. + Decent.}     & &  99.15 & 0.943 & \textbf{468.69} & & 97.34 & 0.949 & \textbf{501.49} & & 88.76 & 0.826 & \textbf{543.43} \\
      & & ($$25.68) & & ($$203.27) & &  ($$26.59) & & ($$229.88) & & ($$28.85) & & ($$249.64) \\
\hline
\textcolor{dd}{Decent. + Decent.} & & 103.72  & 0.985 & 501.96  & & 104.69 & \textbf{0.993} & 528.25 & & 96.71 & 0.9 & 580.76 \\
      & & ($20.81$)  & &  ($201.18$)  & &  ($19.64$) & & ($209.52$) & & ($25.35$) & & ($240.11$) \\
\hline
\end{tabular}
}
\end{table}

Table~\ref{tab:task1_deployment} shows the results from the $100$ evaluation runs on different terrains for the different architectures after completion of training.
All trained architectures achieved high success ratios even on hard, bumpy terrains not seen during training. In other words, the studied architectures all are robust in different unseen terrain and still solve the task. With respect to power consumption, decentralized architectures had a much lower power consumption and show an advantage compared to the centralized approaches. But inside the group of decentralized approaches, introducing furthermore an hierarchical organization showed to be further beneficial. For the flat decentralized architecture (\textcolor{dec}{1\_Dec}) the power consumption was higher than that of both hierarchical architectures with a decentralized lower level (\textcolor{cd}{1\_C\_D} and \textcolor{dd}{1\_D\_D}). This reflects how energy efficient the agent is when combining a hierarchical with a decentralized architecture.

\begin{figure}[!tbh]
    \centering
    \subfloat[\centering ]{{\includegraphics[scale=0.25]{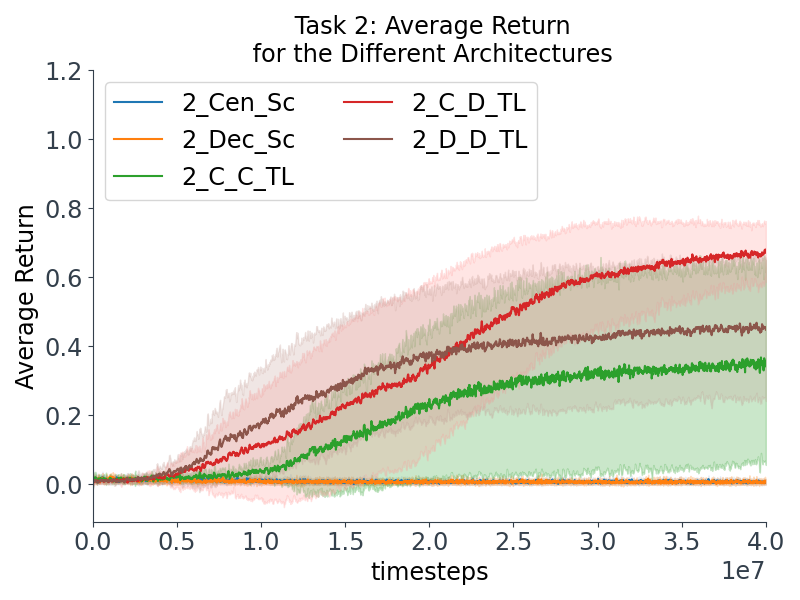} }}%
    \qquad
    \subfloat[\centering ]{{\includegraphics[scale=0.25]{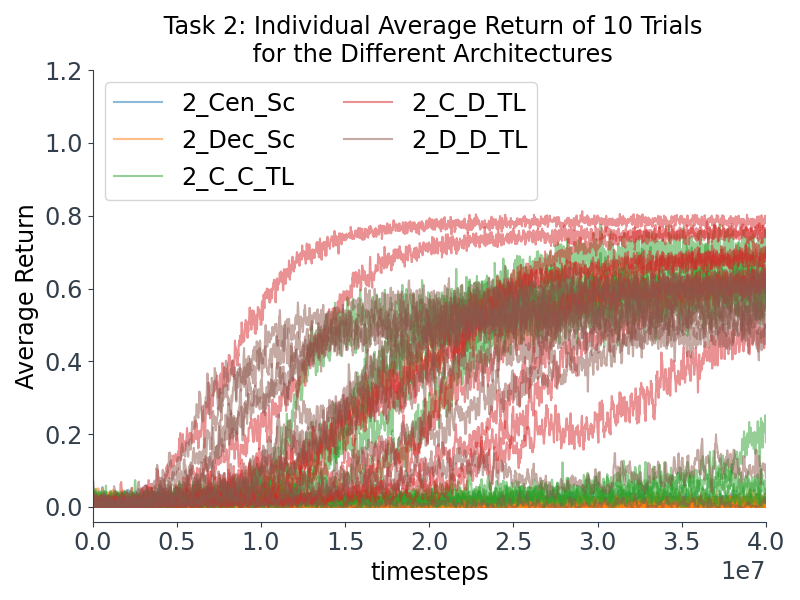} }}%
    \caption[Task 2: Average return of ten trials for the different architectures.]{Average return of ten trials for the different architectures during learning in task 2. (a) The average return of all ten trials. The shaded area represents the limits of the standard deviation and the plot represents the mean value. (b) Visualizing return for all ten individual seeds for the different architectures in task 2 during training.}%
    \label{fig:task2_averageReward}%
\end{figure}

\subsection{Second Task: Seek Target}
\begin{figure}[!tbh]
    \centering
    \includegraphics[scale=0.3]{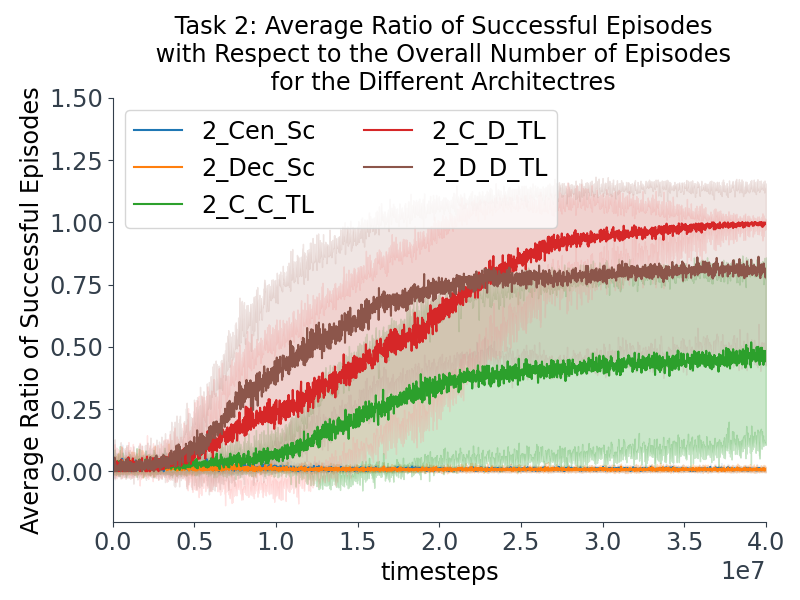}
    \caption[Task 2: Visualization of the ratio of successful episodes with respect to the overall number of episodes for the different architectures.]{Visualization of the ratio of successful episodes with respect to the overall number of episodes for the different architectures in task 2 during training.}
\label{fig:task2_successEpisode}     
\end{figure}

In the second experiment, the agent was aiming for reaching a target, but received only a sparse reward when this was reached. In this case, all flat (non hierarchical) architectures were trained from scratch (indicated in the figures and result tables by the suffix \_SC). In contrast, hierarchical architectures used LLC with frozen parameters that were pretrained on task 1. Only the randomly initialized HLC was trained (as done in transfer learning, indicated in the results by the suffix \_TL). This assumes that during training in task 1 on random goals the agent learns to walk in the environment and that this ability is retained now in task 2, but now the agent has to search for the new target position while still being able to reuse already acquired locomotion skills in the lower level. Fig. \ref{fig:task2_averageReward} shows that policies trained from scratch did not learn the task properly as should be expected (see as well Fig. \ref{fig:task2_successEpisode}). When only using a sparse reward, the task became too difficult. However, using a pretrained LLC in the case of the three hierarchical architectures showed much higher returns. The \textcolor{cd}{2\_C\_D\_TL} architecture outperformed the other hierarchical architectures, while the hierarchical, both level centralized architecture (\textcolor{cc}{2\_C\_C\_TL}) trained the slowest.



\begin{table}[!tbh]
\caption[Task 2: Results of the trained controllers in the deployment phase.]{Task 2: Results for the trained controllers in the deployment phase for controllers trained from scratch and others using transfer learning. The values represent the mean value and the standard deviation (in parenthesis) over ten trials. Each trial is tested over 100 episodes. The best achieved values are highlighted. }
\label{tab:task2_deployment}  
\resizebox{\linewidth}{!}{%
\begin{tabular}{p{2.7cm}@{\extracolsep{8mm}}c@{\extracolsep{4mm}}c@{\extracolsep{4mm}}c@{\extracolsep{10mm}}c@{\extracolsep{4mm}}c@{\extracolsep{4mm}}c@{\extracolsep{10mm}}c@{\extracolsep{4mm}}c@{\extracolsep{4mm}}c}
\hline\noalign{}
Architecture & \multicolumn{3}{c}{Flat terrain (1.0)} & \multicolumn{3}{c}{Uneven terrain (0.8)} & \multicolumn{3}{c}{Bumpy terrain (0.6)} \\
\cline{2-4}\cline{5-7}\cline{8-10}\noalign{}
(HLC + LLC) & Return & Ratio & $P_{suc.}$ & Return & Ratio & $P_{suc.}$ & Return & Ratio & $P_{suc.}$  \\
\noalign{}\hline
\hline
\multicolumn{10}{c}{From scratch} \\
\hline\noalign{}\hline
\textcolor{cen}{Centralized}      & 0.0 & 0.005 & 4433.46 & 0.01 & 0.01 & 4678.03 & 0.01 & 0.009 & 3477.72 \\
 &(0.06) & & (4825.27)  & (0.09) & & (4183.50) &(0.09) & & (1726.77)  \\
\hline
\textcolor{dec}{Decentralized}       & 0.01 & 0.008 & 7995.00& 0.01 & 0.006 & 3927.36 & 0.01 & 0.009 & 3067.37 \\
\noalign{}\hline
\hline
\multicolumn{10}{c}{Transfer learning (pre-trained on random targets in task 1)} \\
\hline\noalign{}\hline
\textcolor{cc}{Cent. + Cent.}      & 0.34 & 0.449 & 2991.25 &0.36 & 0.475 & 3081.00 & 0.27 & 0.354 & 3049.25  \\
      & (0.39) & & (1963.93)& (0.39) & & (2017.79) &(0.37) & & (1949.27)  \\
\hline
\textcolor{cd}{Cent. + Decent.}       & \textbf{0.67} & \textbf{0.996} & \textbf{1448.00} & \textbf{0.64} & \textbf{0.985} & \textbf{1512.89} & \textbf{0.45} & \textbf{0.754} & \textbf{1698.74} \\
   & (0.16) & & (700.28) &(0.19) & & (736.02)  & (0.32) & & (892.17) \\
\hline
\textcolor{dd}{Decent. + Decent.}       & 0.46 & 0.821 & 1924.18 & 0.4 & 0.765 & 2091.21 & 0.26 & 0.543 & 2236.82\\
   &  (0.28) & & (861.37) &(0.29) & & (930.43)  &(0.29) & & (1014.62)  \\
\hline
\noalign{}\hline
\end{tabular}
}
\end{table}

Table~\ref{tab:task2_deployment} again provides the results from 100 evaluation runs on different terrains after training was finished. This illustrates the importance of transfer learning in such sparse tasks. All the flat architectures that were trained from scratch failed the task even after 40 million training steps. However, much improved results and positive return values were observed for the hierarchical architectures that used pre-trained weights from previous tasks. Interestingly, the hierarchical architecture with a centralized HLC and decentralized LLCs (\textcolor{cd}{2\_C\_D\_TL}) achieved the high success ratios. In addition, the values from Table~\ref{tab:task2_deployment} for this architecture demonstrate robustness towards uneven terrain. 

Decentralized low-level hierarchical architectures (\textcolor{cd}{2\_C\_D\_TL} and \textcolor{dd}{2\_D\_D\_TL}) again showed lower power consumption $P_{suc.}$ in successful episodes. In these architectures, the agent did not produce high torques in its joints and was more energy efficient compared to agents with centralized HLCs and LLCs (\textcolor{cc}{2\_C\_C\_TL}. Overall, introducing the decentralized LLCs in a hierarchical architecture using transfer learning boosted performance (higher return values), increased robustness (higher success ratios) and enhanced the energy efficiency of the agent (lower power consumption values). 

\subsection{Third Task: Seek Target in Maze}
In the third task, the agent should navigate through a maze in order to collect a final reward at the end of the maze. As finding a way through the whole maze was too difficult for a random, exploration based search, two subgoals were introduced guiding the agent towards the goal. 
\begin{figure}[!tbh]
    \centering
    \subfloat[\centering ]{{\includegraphics[scale=0.25]{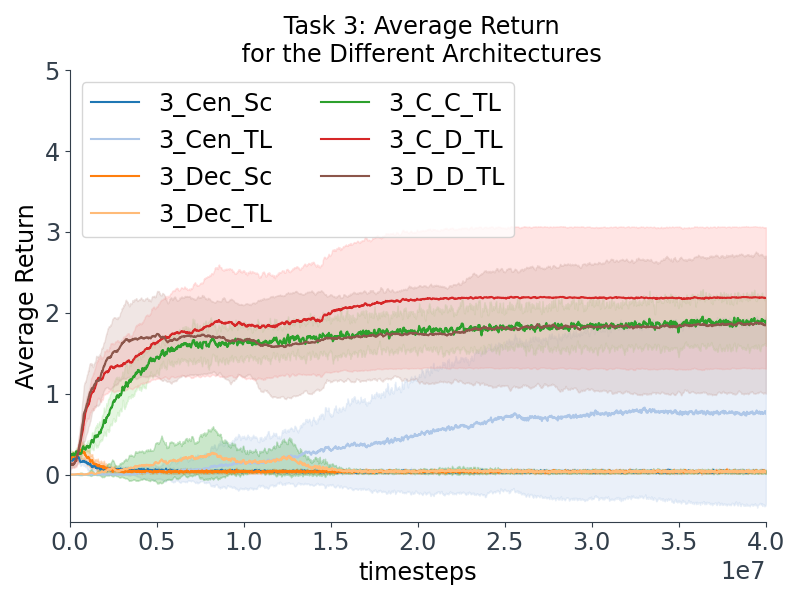} }}%
    \qquad
    \subfloat[\centering ]{{\includegraphics[scale=0.25]{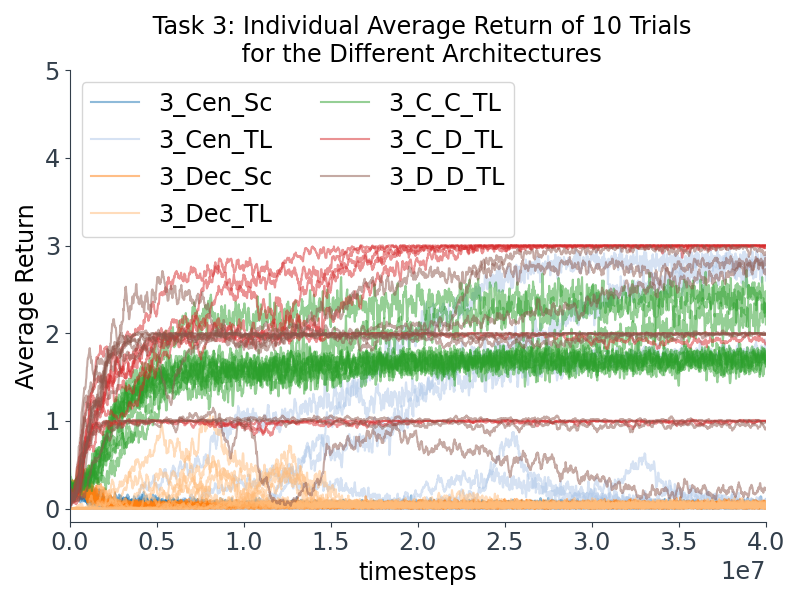} }}%
    \caption[Task 3: Average return of ten trials for the different architectures.]{Average return of ten trials for the different architectures. (a): The average return of all ten trials. The shaded area represents the limits of the standard deviation and the plot represents the mean value. (b): Visualization of the average return of each of the ten seeds trained for every architecture.}%
    \label{fig:task3_averageReward}%
\end{figure}
\begin{figure}[!tbh]
    \centering
    \includegraphics[scale=0.25]{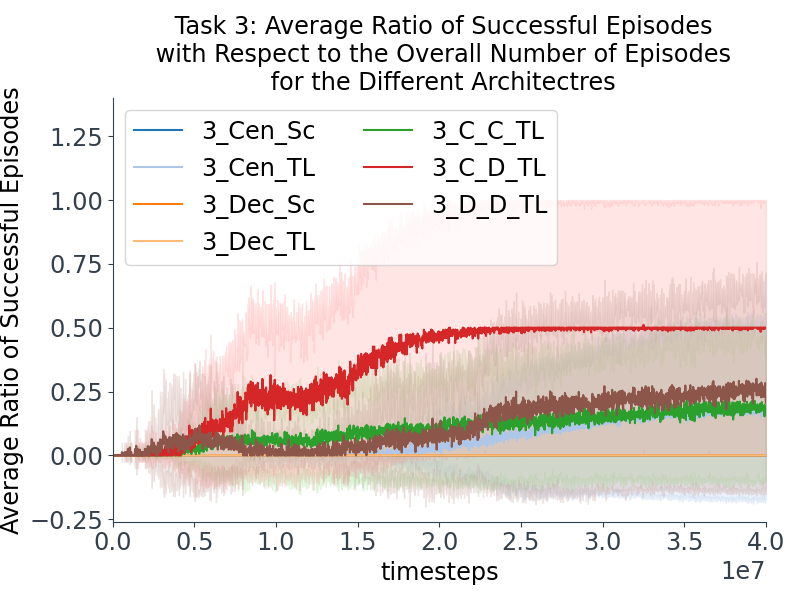}
    \caption[Task 3: Visualization of the ratio of successful episodes with respect to the overall number of episodes for the different architectures.]{Visualization of the ratio of successful episodes with respect to the overall number of episodes for the different architectures.}
\label{fig:task3_successEpisode}     
\end{figure}
Fig.~\ref{fig:task3_averageReward} and \ref{fig:task3_successEpisode} show results during training. As the task requires memorizing a path through the maze, the neural network architecture was extended by an additional layer of LSTM units (see sect. \ref{subsect:setup}). In this task, flat architectures were trained either from scratch (\textcolor{cen}{3\_Cen\_Sc}, \textcolor{dec}{3\_Dec\_Sc}) or---following the sequential protocol as used in transfer learning approaches---were pretrained for task 1 and shown are results for subsequential training in the maze environment (\textcolor{cenTL}{3\_Cen\_TL}, \textcolor{decTL}{3\_Dec\_TL}). All these non-hierarchical approaches showed little to no success. In contrast, hierarchical architectures again showed transfer learning capabilities and were able to collect a considerable return. In particular, agents with a centralized HLC and decentralized LLCs (\textcolor{cd}{3\_C\_D\_TL}) learned faster and reached higher return values as they were able to solve the maze at a considerable rate (Fig. \ref{fig:task3_successEpisode}).


\begin{table}[!tbh]
\caption[Task 3: Results of the trained controllers in the deployment phase.]{Results of the trained controllers for task 3 in the evaluation phase. 
Values represent the mean value and the standard deviation (in parenthesis)  over 100 episodes and for ten different training seeds. The best values are highlighted. }
\label{tab:task3_deployment}  
\resizebox{\linewidth}{!}{%
\begin{tabular}{p{2.7cm}@{\extracolsep{8mm}}c@{\extracolsep{4mm}}c@{\extracolsep{4mm}}c@{\extracolsep{10mm}}c@{\extracolsep{4mm}}c@{\extracolsep{4mm}}c@{\extracolsep{10mm}}c@{\extracolsep{4mm}}c@{\extracolsep{4mm}}c}
\hline\noalign{}
Architecture & \multicolumn{3}{c}{Flat terrain (1.0)} & \multicolumn{3}{c}{Uneven terrain (0.8)} & \multicolumn{3}{c}{Bumpy terrain (0.6)} \\
\cline{2-4}\cline{5-7}\cline{8-10}\noalign{}
(HLC + LLC) & Return & Ratio & $P_{suc.}$ & Return & Ratio & $P_{suc.}$ & Return & Ratio & $P_{suc.}$  \\
\noalign{}\hline
\hline
\multicolumn{10}{c}{From scratch} \\
\hline\noalign{}\hline
\textcolor{cen}{Centralized}      & 0.03 & 0.000 & - & 0.03 & 0.000 & - & 0.02 & 0.000 & - \\
 & (0.19) & & (-) & (0.18) & & (-) & (0.15) & & (-)  \\
\hline
\textcolor{dec}{Decentralized}    &  0.03 & 0.0 & - & 0.02 & 0.0 & -  & 0.02 & 0.0 &  - \\
 &(0.17) & & (-)&(0.13) & & (-) &(0.15) & & (-)   \\
\noalign{}\hline
\hline
\multicolumn{10}{c}{Transfer learning (Random pre-trained trial trained in task 1)} \\
\hline\noalign{}\hline
\textcolor{cenTL}{Centralized}       &  0.76 & 0.17 & 3297.79 & 0.76 & 0.17 & 3538.87 &  0.45 & 0.078 & 4047.24 \\
 &(1.18) & & (1609.73) &(1.18) & & (896.04)  &(0.94) & & (1156.85)    \\
\hline
\textcolor{decTL}{Decentralized}      &  0.03 & 0.0 & - & 0.03 & 0.0 & -  & 0.03 & 0.0 &  - \\
 &(0.17) & & (-) &(0.17) & & (-)  &(0.16) & & (-)   \\
\hline\noalign{}\hline
\textcolor{cc}{Cent. + Cent.}       & 1.87 & 0.186 & 6000.01  & 1.87 & 0.18 & 5847.41 & \textbf{1.27} & 0.099 & 6181.09 \\
 & (0.81) & & (1808.74)   & (0.81) & & (1467.75)  &  (1.01) & & (1399.8) \\
\hline
\textcolor{cd}{Cent. + Decent.}       & \textbf{2.19} & \textbf{0.5} & \textbf{2916.61} & \textbf{2.15} & \textbf{0.49} & \textbf{3169.24} &  \textbf{1.27} & \textbf{0.227} & \textbf{3879.17}  \\
& (0.88) & & (590.8)  & (0.93) & & (699.05)   & (1.15) & & (978.29)  \\
\hline
\textcolor{dd}{Decent. + Decent.}    & 1.86 & 0.257 & 4367.02  &  1.79 & 0.232 & 4698.57 & 1.01 & 0.086 & 4911.36 \\
 &(0.9) & & (962.67)    & (0.92) & & (1078.28)   &(0.98) & & (942.24)  \\
\hline
\noalign{}\hline
\end{tabular}
}
\end{table}

Table~\ref{tab:task3_deployment} shows the results from the 100 evaluation trials for different terrains. The results again support the hypothesis that using decentralized and hierarchical architectures is also advantageous for such complex tasks that require forms of memory. The different hierarchical architectures maintain a level of success for generalization to uneven terrain (smoothness 0.8). But this drops considerably when turning towards bumpy terrain (smoothness equal to 0.6).
The power consumption values $P_{suc.}$ show again that agents with decentralized hierarchical architectures had lower power consumption 
which supports the observations already encountered in the previous tasks. 

\section{Conclusion}
This study investigated the combination of decentralized and hierarchical control architectures for learning of locomotion which was compared to state-of-the-art architectures as the standard centralized architecture or a simple hierarchical approach. Overall, it showed that adding concurrent decentralized modules across different vertical levels of a control hierarchy facilitates the learning process and ensures more robustness as well as more energy efficient behavior. Furthermore, the advantage of hierarchical architectures for transfer learning was maintained. 
The results demonstrate that the horizontal modularity and vertical temporal abstraction can be used together to improve modern architectures and solve more goal-directed tasks that require transfer learning. This facilitates faster learning, avoids catastrophic forgetting, and allows to reuse learned skills. 
This study provided a starting point for combination of such modular organizations of control approaches.
Future research should consider fine-tuning the networks' architectures, use more layers, or change the degree of decentralization. 
Another point worth investigating is the robustness of these methods against specific leg failures after training, and whether the trained HLCs can adapt to such disturbances. This can be further explored to test the limits of such architectures. It should be noted that this work refers to a simulated robot. Therefore, it would be also interesting to test these architectures on real robots. 
%
%
%
\bibliographystyle{splncs04}

%
\bibliography{references}

\end{document}